%% file: main.tex
\documentclass[10pt,twocolumn,letterpaper]{article}

\usepackage[pagenumbers]{cvpr} 
\usepackage{fancyhdr}
\input{preamble}

\usepackage{bm} 
\definecolor{cvprblue}{rgb}{0.21,0.49,0.74}
\usepackage[pagebackref,breaklinks,colorlinks,citecolor=cvprblue]{hyperref}
\usepackage{lipsum} 

\def\confName{CVPR}
\def\confYear{2025}

\title{V-NAW: Video-based Noise-aware Adaptive Weighting \\ for Facial Expression Recognition}

\author{JunGyu Lee\textsuperscript{1,2}, Kunyoung Lee\textsuperscript{1}, Haesol Park\textsuperscript{1}, Ig-Jae Kim\textsuperscript{1,2}
Gi Pyo Nam\textsuperscript{1,2}
\\
\textsuperscript{1} Korea Institute of Science and Technology, Seoul, Korea\\ \textsuperscript{2} AI-Robotics, KIST School, University of Science and Technology, Daejeon, Korea\\
{\tt\small \{jungyu0413, kyle, haesol, drjay, gpnam\}@kist.re.kr}
}

\begin{document}
\maketitle

\input{sec/0_abstract}    
\input{sec/1_intro}

\input{sec/2_related_work}
\input{sec/3_method}
\input{sec/4_experiment}

\input{sec/5_conclusion}

{
    \small
    \bibliographystyle{ieeenat_fullname}
    \bibliography{main}
}


\end{document}

%% file: preamble.tex
\usepackage[dvipsnames]{xcolor}


%% file: sec/0_abstract.tex
\begin{abstract}
Facial Expression Recognition (FER) plays a crucial role in human affective analysis and has been widely applied in computer vision tasks such as human-computer interaction and psychological assessment.
The 8th Affective Behavior Analysis in-the-Wild (ABAW) Challenge aims to assess human emotions using the video-based Aff-Wild2 dataset.
This challenge includes various tasks, including the video-based EXPR recognition track, which is our primary focus.
In this paper, we demonstrate that addressing label ambiguity and class imbalance, which are known to cause performance degradation, can lead to meaningful performance improvements.
Specifically, we propose Video-based Noise-aware Adaptive Weighting (V-NAW), which adaptively assigns importance to each frame in a clip to address label ambiguity and effectively capture temporal variations in facial expressions.
Furthermore, we introduce a simple and effective augmentation strategy to reduce redundancy between consecutive frames, which is a primary cause of overfitting.
Through extensive experiments, we validate the effectiveness of our approach, demonstrating significant improvements in video-based FER performance.
\end{abstract}

%% file: sec/1_intro.tex
\section{Introduction}
\label{sec:intro}
Facial Expression Recognition (FER) has been widely applied in various fields, such as human-robot interaction, non-verbal communication, and mental health monitoring.
Previous studies have introduced various methods to address key challenges in facial expression recognition, such as occlusion~\cite{lee2023latent, wang2020region}, label ambiguity~\cite{IPA2LT, EAC, SCN, DMUE, RUL, NLA}, and class imbalance~\cite{face2exp, MEK, NLA}.
Some studies have focused on constructing datasets, including those collected in controlled laboratory environments~\cite{ckplus, pantic2005web_mimi, lyons1998coding_JAFFE}, as well as those captured in in-the-wild settings~\cite{ferlpus, rafdb, affectnet}.
However, most approaches rely on image datasets, limiting their ability to capture the temporal variations of facial expressions. As human perception inherently relies on both spatial and temporal cues, capturing temporal variations is crucial for accurate emotion recognition. 

The Affective Behavior Analysis in-the-Wild (ABAW) Challenge has been organized to analyze human affect, utilizing the Aff-Wild2~\cite{kollias2018aff} dataset, a large-scale video dataset collected in unconstrained environments.
In contrast to traditional FER benchmarks that primarily focus on images, Aff-Wild2 provides images in a video format, enabling a more comprehensive analysis of temporal variations in facial expressions in real-world scenarios.

In FER, the main challenges include label ambiguity, as facial expressions are inherently subjective and subject to annotator interpretation, and class imbalance, where some expressions are overrepresented in daily life while others are less frequent. These factors contribute to model uncertainty and bias toward majority classes. Previous image-based FER studies~\cite{DMUE, SCN, LANet, EAC, face2exp, MEK, NLA} have tried to address these issues, however, these methods perform FER on individual frames, without considering the temporal variations in facial expressions.
In addition to these challenges, video datasets often suffer from spatiotemporal redundancy, where consecutive frames contain highly similar information. This redundancy can cause models to overfit on specific regions or redundant facial patterns rather than capturing meaningful temporal variations. To alleviate the problem, several data augmentation approaches have been proposed~\cite{tong2022videomae, cai2023marlin}, such as clip skipping and frame erasing.

To address these challenges at the same time, we propose Video-based Noise-aware Adaptive Weighting (V-NAW), an extension of Navigating Label Ambiguity (NLA)~\cite{NLA} for video-based FER. V-NAW adaptively assigns weights to each sample, effectively mitigating label ambiguity and class imbalance in video-based FER datasets.
Additionally, we introduce an augmentation strategy using frame skipping and in-frame erasing to mitigate spatiotemporal redundancy of the video dataset. 
This enables the model to learn meaningful temporal information rather than overfitting to specific regions or repetitive patterns.
Through extensive experiments on the Aff-Wild2 dataset, we demonstrate that our method significantly outperforms the baseline, highlighting its effectiveness for video-based FER.

The main contributions of our work are as follows:
\begin{itemize}
\item We propose Video-based Noise-aware Adaptive Weighting (V-NAW), an extension of NAW for video-based FER. V-NAW adaptively assigns loss weights based on the ambiguity of entire video clips, enhancing robustness against label noise and class imbalance.
\item We introduce an augmentation strategy to mitigate data redundancy in video-based FER. This includes frame skipping to remove redundant frames and in-frame erasing to mask some of the pixels, preventing the model from overfitting to specific regions or repetitive patterns.
\item We conduct extensive experiments on the Aff-Wild2 dataset, demonstrating the effectiveness of our method with superior performance.
\end{itemize}

%% file: sec/2_related_work.tex
\section{Related Work}
\label{sec:relate}
Facial Expression Recognition (FER) can be categorized into image-based and video-based approaches. Recent studies have highlighted the challenges of FER in unconstrained environments, where variations in facial expressions, occlusions, and subjective annotations lead to significant discrepancies \cite{baseline2025, abaw7, abaw6, kollias2022abaw, kollias2023abaw2, kollias2023abaw, kollias2024distribution, kollias2021distribution, kollias2019face}.
In addition, image-based approaches struggle to capture the temporal variations of facial expressions. To overcome this limitation, recent research trends have increasingly focused on video-based FER, leveraging sequential information to enhance expression recognition in real-world scenarios.

\subsection{Image-based FER}
Facial Expression Recognition (FER) encounters two key challenges: (1) noisy labels, which arise due to the inherent ambiguity of facial expressions and inconsistencies in human annotations, and (2) class imbalance, where certain expressions appear more frequently than others, leading to biased model predictions.
To mitigate the impact of noisy labels, various methods have been proposed. SCN~\cite{SCN} employs re-labeling and ranking regularization to suppress label noise, while DMUE~\cite{DMUE} explores latent label distributions using multiple branches and uncertainty estimation. RUL~\cite{RUL} leverages feature mixup and a multi-branch framework to learn from relative sample difficulty, and MAN~\cite{zhang2022man} introduces a co-division module that separates clean and noisy samples to enhance discriminative ability. Lastly, EAC~\cite{EAC} ensures model robustness by aligning Class Activation Maps (CAMs) \cite{zhou2016learning_CAM} across original and flipped images, preventing models from overfitting to misleading features. 
Class imbalance has also been a critical challenge in FER, as imbalanced datasets often result in models biased toward majority classes. Face2Exp~\cite{face2exp} integrates large-scale unlabeled face recognition data through a meta-optimization framework. Zhang et al~\cite{MEK}, in particular, re-balances attention consistency and applies label smoothing to further improve minority class representations.
Instead of prior approaches that address either label ambiguity or class imbalance separately, NLA~\cite{NLA} is the first method to tackle both challenges simultaneously by assigning adaptive weighting based on each sample's uncertainty.

\subsection{Video-based FER}  
Video-based FER methods~\cite{CtyunAI2024, hsemotion2024} extend image-based FER by leveraging the temporal dynamics of facial expressions. Instead of analyzing a single frame in isolation, these approaches learn from sequences of frames to capture how expressions evolve over time.
SAANet~\cite{SAANet} utilizes LSTM~\cite{graves2012long} networks and related gated RNN~\cite{grossberg2013recurrent} variants, which have proven effective in modeling subtle changes in facial appearance over consecutive frames. MARLIN~\cite{cai2023marlin} is a self-supervised Masked AutoEncoder (MAE)~\cite{he2022masked} that learns universal facial representations from videos by reconstructing spatiotemporal facial details. Lin et al.~\cite{M2-Lab-Purdue2024} propose a lightweight framework for expression classification and AU detection using a frozen CLIP encoder with a trainable MLP. Yu et al.~\cite{USTC-IAT-United2024} tackle dataset limitations and class imbalance in FER by leveraging semi-supervised learning for pseudo-labeling and implementing a debiased feedback learning strategy. Zhang et al.~\cite{netease2024} introduce a Transformer-based multi-modal approach, employing a MAE for facial feature extraction and a scene-specific classifier, achieving strong performance in ABAW challenges~\cite{abaw7, abaw6}.

Despite these advancements, extracting temporal dynamics in FER remains a significant challenge. In previous ABAW challenges, various methods~\cite{kollias2023multi, kollias2021analysing, kollias2021affect, kollias2020analysing, kollias2019expression, kollias2019deep, zafeiriou2017aff} have been proposed to enhance the learning of temporal variations. While LSTM-based methods effectively capture short-term dependencies, they struggle with long-range relationships, making them less suitable for video-based FER. To address this limitation, Transformer-based approaches have been increasingly adopted in recent ABAW challenges~\cite{abaw6}, proving to be more effective in capturing global temporal dependencies.

Inspired by NLA~\cite{NLA}, which addresses both label ambiguity and class imbalance simultaneously, we extend it to video-based FER by incorporating a Transformer encoder, following previous works \cite{USTC-IAT-United2024, CtyunAI2024, netease2024}, to effectively learn temporal features.

\begin{figure*}[t]
\begin{center}
\includegraphics[width=1.0\linewidth]{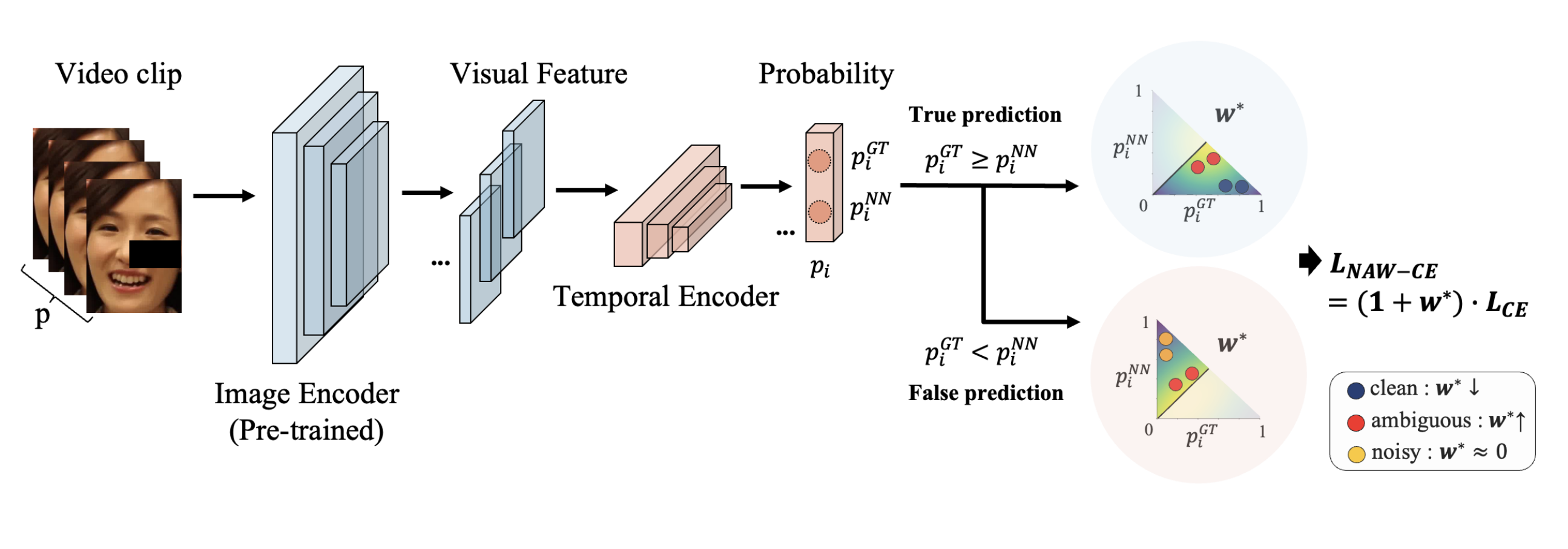}
\end{center}
\vspace{-2em}
\caption{The overview of our proposed method. First, we apply augmentation to the input video clip. Next, we extract frame-wise visual features using a pre-trained image encoder (blue box). These features are then aggregated at the clip level and processed by a temporal encoder to capture temporal information (pink box). Finally, we incorporate Noise-aware Adaptive Weighting (NAW)~\cite{NLA}}
\label{fig:pipeline}
\vspace{-1.5em}
\end{figure*}

%% file: sec/3_method.tex
\section{Method}
\label{sec:method}
\subsection{Overview}
In this section, we describe our method for video-based FER, which consists of three main stages and is illustrated in Figure~\ref{fig:pipeline}. First, we apply data augmentation techniques, including frame skipping and pixel erasing, to mitigate redundancy and enhance model robustness. Second, an image encoder extracts frame-wise features, which are then processed by a temporal encoder to capture sequential dependencies. Finally, Noise-aware Adaptive Weighting (NAW) adaptively adjusts the loss function based on sample ambiguity, enhancing generalization to noisy labels. Further details on each stage are provided in the following subsections.

\subsection{Data Augmentation}
Since consecutive video frames often contain highly similar information, we apply an augmentation strategy for video-based FER to mitigate spatiotemporal redundancy and enhance model generalization. Specifically, we introduce two techniques: (1) frame skipping, which randomly removes entire frames within a video clip to reduce temporal redundancy, and (2) in-frame erasing, which randomly masks 20\% of pixels in each frame to prevent the model from overfitting to specific facial regions. 

To ensure clarity in the following subsection, we first define the notations used in our method. A dataset $\mathcal{D}$ consists of $N$ video clips, where each clip contains $P$ frames. We define $\mathcal{D}$ as:
\begin{equation}
    \mathcal{D} = \{(X_i, y_i)\}_{i=1}^{N},
\end{equation}
where $X_i = \{x_i^p\}_{p=1}^{P}$ represents the $i$-th video clip composed of $P$ frames, and $x_i^p$ denotes the $p$-th frame. The corresponding ground-truth label for the entire clip is given as:
\begin{equation}
    y_i \in \{1, \cdots, K\}.
\end{equation}
These notations will be used throughout the subsequent subsections.

\subsection{Frame Skipping}
Given a video clip $X_i = \{x_i^p\}_{p=1}^{P}$ consisting of $P$ frames, we define a binary masking function $M_i^p \sim \text{Bernoulli}(1 - \lambda)$ to determine whether the $p$-th frame is retained ($M_i^p = 1$) or erased ($M_i^p = 0$). The parameter $\lambda \in [0,1]$ controls the proportion of removed frames. For example, if $\lambda$ is randomly sampled from the range $[0.2, 0.4]$, it means that between 20\% and 40\% of the frames in each video clip are randomly dropped. The frame-skipped video clip is then represented as:
\begin{equation}
    \tilde{X}_i = \{x_i^p \mid M_i^p = 1\}.
\end{equation}
By randomly skipping frames, the model is encouraged to learn more robust temporal representations rather than overfitting to redundant sequential patterns.

\subsection{In-frame Random Erasing}
To further prevent overfitting to specific facial regions, we apply a random patch erasing strategy to each retained frame. Specifically, for each video clip $\tilde{X}_i = \{x_i^p \mid M_i^p = 1\}$, we randomly select a set of pixels $\Omega_i^p$ from the set of all pixel locations $\mathcal{U}$ such that:
\begin{equation}
    |\Omega_i^p| = 0.2 \times |\tilde{x}_i^p|.
\end{equation}
This ensures that exactly 20\% of the pixels are randomly erased in each frame. The same set of pixels $\Omega_i^p$ is masked across all frames in the clip. The final augmented video clip is then given by:
\begin{equation}
    \hat{X}_i = \{M_i^p \cdot (\tilde{x}_i^p \odot \mathbb{1}_{\Omega_i^p})\}_{p=1}^{P},
\end{equation}
where $\mathbb{1}_{\Omega_i^p}$ is a binary mask that is 0 for the selected 20\% of pixels and 1 elsewhere, and $\odot$ represents element-wise multiplication.

\begin{table*}[!ht]
\renewcommand\arraystretch{1.05} 
\centering
  \caption{F1-score results (in \%) for models trained and tested on different folds with a 7:3 train-to-test ratio, including the original training/validation set of the \textit{Aff-Wild2} dataset.}
  \label{tab:exp_F1_val}
  \vspace{-0.7em}

\resizebox{\linewidth}{!}{%
\setlength{\tabcolsep}{0.7 em}
\renewcommand\arraystretch{1.2}
  \begin{tabular}{lccccccccc}
    \hline
    \textbf{Val Set} & \textbf{Neutral} & \textbf{Anger} & \textbf{Disgust} & \textbf{Fear} & \textbf{Happiness} & \textbf{Sadness} & \textbf{Surprise} & \textbf{Other} & \textbf{Avg.} \\[1pt]
    \hline \hline
    Official &  57.12 & 45.76 & 33.45 & 23.34 & 43.67 & 50.12 & 30.23 & 50.82 &  41.81 \\
    fold-1  &  55.98 &  35.12 &  24.76 &  20.45 &  55.23 &  54.34 &  34.45 &  40.89 &  40.15 \\
    fold-2  &  50.34 &  41.89 &  23.45 &  22.76 &  47.12 &  52.98 &  32.78 &  42.22 &  39.19 \\
    fold-3  &  52.45 &  38.12 &  23.34 &  22.98 &  38.78 &  54.45 &  33.67 &  47.12 &  38.86 \\
    \hline
  \end{tabular}}
\end{table*}

\begin{table*}[ht]
\renewcommand\arraystretch{1.05} 
\centering
  \caption{Ablation Study Results on Aff-Wild2. This table demonstrates how our method enhances discriminative ability by adaptively assigning weights to each sample through Augmentation (Aug) and Noise-aware Adaptive Weighting (NAW).}
  \label{tab:ablation}
  \vspace{-0.7em}

\resizebox{\linewidth}{!}{%
\setlength{\tabcolsep}{0.7 em}
\renewcommand\arraystretch{1.2}
  \begin{tabular}{ccccccccccc}
    \hline
    \textbf{Aug} & \textbf{NAW} &\textbf{Neutral} &\textbf{Anger} &\textbf{Disgust} &\textbf{Fear} &\textbf{Happiness} &\textbf{Sadness} &\textbf{Surprise} &\textbf{Other} &\textbf{Avg.} \\[1pt]
    \hline \hline
     &  &  48.67 & 42.46 & 27.44 & 17.26 & 42.10 & 47.16 & 22.23 & 45.96 & 36.66 \\
    \checkmark &  &  52.56 & 42.11 & 26.66 & 17.93 & 43.73 & 48.99 & 24.25 & 47.64 & 37.98 \\
    & \checkmark  &  50.12 & 45.67 & 34.89 & 19.78 & 45.02 & 50.13 & 29.10 & 47.34 & 40.26 \\
    \checkmark  & \checkmark & 57.12 & 45.76 & 33.45 & 23.34 & 43.67 & 50.12 & 30.23 & 50.82 &  41.81 \\
    \hline
  \end{tabular}}
\end{table*}

\subsection{Image Encoder}
To obtain spatial-level representations, we first construct an image-based FER dataset by combining three widely used datasets: FERPlus~\cite{ferlpus}, RAF-DB~\cite{rafdb}, and AffectNet~\cite{affectnet}. We pretrain an image encoder on a classification task utilizing NLA~\cite{NLA}. The feature extractor from the pre-trained model is then utilized as the image encoder, which processes each frame of a video clip independently to extract visual features. Given a video clip $\hat{X}_i = \{\hat{x}_i^p\}_{p=1}^{P}$ consisting of $P$ frames, we extract the frame-level visual features as follows:
\begin{equation}
    F_v = \{F_v^p\}_{p=1}^{P}, \quad F_v^p = \text{Image Encoder}(\hat{x}_i^p).
\end{equation}
By leveraging this pretrained image encoder, we ensure that each frame is represented with high-quality feature embeddings.

\subsection{Temporal Encoder}
To capture temporal feature across video frames, we utilize a Transformer encoder. This module processes the sequence of extracted visual features and extracts their temporal information. The temporal features are computed as:
\begin{equation}
H_i = \text{Transformer Encoder}(F_v),
\end{equation}
where $H_i = \{h_i^p\}_{p=1}^{P}$ represents the temporally encoded features for the video clip. Next, a fully connected layer followed by a softmax function is applied to extract frame-wise class probabilities:
\begin{equation}
P_i = \{p_{i,p}\}_{p=1}^{P}, \quad p_{i,p}^k = \frac{e^{h_{i,p}^k}}{\sum_{j=1}^{K} e^{h_{i,p}^j}},
\end{equation}
where $p_{i,p}^k$ denotes the probability of class $k$ for the $p$-th frame. Finally, to obtain the clip-level prediction, we aggregate the frame-wise probabilities across all $P$ frames as:
\begin{equation}
    p_i^k = \frac{1}{P} \sum_{p=1}^{P} p_{i,p}^k.
\end{equation}
\subsection{Video-based Noise-aware Adaptive Weighting}

After obtaining the probabilities, we apply the Video-based Noise-aware Adaptive Weighting (V-NAW) to adaptively assign weights to the loss function based on prediction uncertainty. NAW assigns sample-specific weights according to ambiguity, using a Gaussian kernel.

\noindent\textbf{Cross-Entropy Loss (CE)}  
Given a video clip $\mathcal{D} = \{(X_i, y_i)\}_{i=1}^{N}$ with class probabilities $p_i^k$, the standard cross-entropy loss is defined as:

\begin{equation}
    \mathcal{L}_{\text{CE}}= -\frac{1}{N} \sum_{i=1}^{N} \log p_i^{y_i}.
\end{equation}

\noindent\textbf{Noise-aware Adaptive Weighting}  
To handle label ambiguity, NAW introduces an adaptive weight computed via a Gaussian kernel:

\begin{equation}
    w^{*}(\mathbf{p}_i| \bm{\mu}, \bm{\Sigma})= C\cdot\exp \left( -\frac{1}{2} (\mathbf{p}_i - \bm{\mu})^T \bm\Sigma^{-1} (\mathbf{p}_i - \bm{\mu}) \right),
\end{equation}

where $\bm{\mu}$ is the mean vector, $\bm{\Sigma}$ is the covariance matrix, and $C=(2\pi \sqrt{|\bm{\Sigma}|})^{-1}$ is the normalizing constant.We follow the parameter settings of NLA~\cite{NLA}. For more details, please refer to the original paper.

\noindent\textbf{Total Loss}  
The total loss for video clips is then computed as:

\begin{equation}
\mathcal{L}_{\text{NAW-CE}} = (1+w^{*}(\mathbf{p}_i)) \cdot \mathcal{L}_{\text{CE}}.
\end{equation}

This video-based NAW enables the model to adaptively adjust its learning process based on the ambiguity of the entire video clip, rather than individual frames, thereby improving robustness in video-based FER tasks. For further details about NAW, please refer to NLA~\cite{NLA}.

%% file: sec/4_experiment.tex
\section{Experiment}
\label{sec:exp}

In this section, we first describe the evaluation metrics and datasets, followed by the implementation details. We then assess our model's performance using the ABAW8 competition metrics.

\subsection{Evaluation metrics}To assess the model performance on each track, ABAW challenge defines a specific evaluation metric for EXPR. The performance is evaluated using the macro-averaged F1 score across all 8 expression categories, which is computed as:

\begin{equation}
F1 = \frac{2 \times {Precision} \times {Recall}}{{Precision} + {Recall}},
\end{equation}

where precision and recall are defined as:

\begin{equation}
{Precision} = \frac{{TP}}{{TP} + {FP}}, \quad {Recall} = \frac{{TP}}{{TP} + {FN}}.
\end{equation}

The final performance score $P$ is obtained by averaging the F1 scores across all categories:

\begin{equation}
    P = \frac{1}{8} \sum_{c=1}^{8} F1_c.
\end{equation}

Here, $c$ represents the category index, $TP$ refers to True Positives, $FP$ to False Positives, and $FN$ to False Negatives.

\subsection{Implementatal Setting}

We utilize RetinaFace~\cite{deng2020retinaface} to detect faces in each frame and normalize them to a resolution of $224 \times 224$. We pretrain an NLA~\cite{NLA} model on a static facial expression dataset, which consists of FERPlus~\cite{ferlpus}, RAF-DB~\cite{rafdb}, and AffectNet~\cite{affectnet}, using a Swin Transformer \cite{liu2021swin} backbone. We then fine-tune the model on the Aff-Wild2 dataset with a batch size of 4096 using 8 NVIDIA A100 GPUs. The video clips used for training contain $P = 100$ frames, and the learning rate is set to $0.0001$ with the AdamW optimizer~\cite{loshchilov2017decoupled}.

\subsection{Results}
\noindent\textbf{Evaluation}  
In this section, we conduct experiments to evaluate the effectiveness of our model in the expression recognition task. We perform k-fold validation with a 7:3 train-to-test ratio. The F1 scores for each fold, along with the average F1 score, are presented in Table~\ref{tab:exp_F1_val}. Our method achieves a significant improvement over the baseline (25.00\%), outperforming it by a margin of 13.86\% in terms of F1-score. Furthermore, NAW addresses class imbalance by assigning the highest weight to ambiguous samples, which contain a higher proportion of minority class. As a result, we observe an improvement in the performance of minority classes compared to the baseline. Importantly, unlike other recent models that leverage additional modalities such as audio or text, our approach relies solely on visual features extracted from each frame. Despite this limitation, our model demonstrates strong performance, highlighting its ability to effectively learn discriminative visual representations without the extra information.

\noindent\textbf{Ablation}  
We also conduct an ablation study to assess the contribution of each component in our model, including the baseline (Vanilla), the Noise-aware Adaptive Weighting (NAW), and the augmentation strategy. Table~\ref{tab:ablation} presents the performance of each strategy. Our results demonstrate the effectiveness of our method compared to the vanilla model, achieving improvements of 1.32\% with Augmentation, 3.6\% with NAW, and 5.15\% when both are applied.

%% file: sec/5_conclusion.tex
\section{Conclusion}
\label{sec:conclusion}
In this paper, we propose a video-based Facial Expression Recognition (FER) method that addresses several key challenges in the field. First, we introduce Video-based Noise-aware Adaptive Weighting (V-NAW), an extension of Noise-aware Adaptive Weighting (NAW), to tackle label ambiguity and noisy label issues. Second, we propose an augmentation strategy to reduce spatiotemporal redundancy, which often leads to model overfitting, thereby enhancing generalization. Furthermore, we incorporate a temporal encoder to effectively capture long-range dependencies in facial expressions across video frames.
Through extensive experiments, we demonstrate that integrating NAW and the augmentation strategy results in substantial performance improvements on the Aff-Wild2 dataset, highlighting the importance of addressing label noise and mitigating spatiotemporal redundancy in video-based FER. Importantly, despite relying exclusively on visual features without incorporating additional modalities such as audio or text, our approach achieves competitive performance against recent multimodal models, highlighting the strength of vision-only methods for facial expression recognition.

\section*{Acknowledgments}
This research was supported by KIST Institutional Program (Project No. 2E33612).

%% file: main.bbl
\begin{thebibliography}{50}
\providecommand{\natexlab}[1]{#1}
\providecommand{\url}[1]{\texttt{#1}}
\expandafter\ifx\csname urlstyle\endcsname\relax
  \providecommand{\doi}[1]{doi: #1}\else
  \providecommand{\doi}{doi: \begingroup \urlstyle{rm}\Url}\fi

\bibitem[Barsoum et~al.(2016)Barsoum, Zhang, Ferrer, and Zhang]{ferlpus}
Emad Barsoum, Cha Zhang, Cristian~Canton Ferrer, and Zhengyou Zhang.
\newblock Training deep networks for facial expression recognition with crowd-sourced label distribution.
\newblock In \emph{Proceedings of the 18th ACM international conference on multimodal interaction}, pages 279--283, 2016.

\bibitem[Cai et~al.(2023)Cai, Ghosh, Stefanov, Dhall, Cai, Rezatofighi, Haffari, and Hayat]{cai2023marlin}
Zhixi Cai, Shreya Ghosh, Kalin Stefanov, Abhinav Dhall, Jianfei Cai, Hamid Rezatofighi, Reza Haffari, and Munawar Hayat.
\newblock Marlin: Masked autoencoder for facial video representation learning.
\newblock In \emph{Proceedings of the IEEE/CVF conference on computer vision and pattern recognition}, pages 1493--1504, 2023.

\bibitem[Deng et~al.(2020)Deng, Guo, Ververas, Kotsia, and Zafeiriou]{deng2020retinaface}
Jiankang Deng, Jia Guo, Evangelos Ververas, Irene Kotsia, and Stefanos Zafeiriou.
\newblock Retinaface: Single-shot multi-level face localisation in the wild.
\newblock In \emph{Proceedings of the IEEE/CVF conference on computer vision and pattern recognition}, pages 5203--5212, 2020.

\bibitem[Graves and Graves(2012)]{graves2012long}
Alex Graves and Alex Graves.
\newblock Long short-term memory.
\newblock \emph{Supervised sequence labelling with recurrent neural networks}, pages 37--45, 2012.

\bibitem[Grossberg(2013)]{grossberg2013recurrent}
Stephen Grossberg.
\newblock Recurrent neural networks.
\newblock \emph{Scholarpedia}, 8\penalty0 (2):\penalty0 1888, 2013.

\bibitem[He et~al.(2022)He, Chen, Xie, Li, Doll{\'a}r, and Girshick]{he2022masked}
Kaiming He, Xinlei Chen, Saining Xie, Yanghao Li, Piotr Doll{\'a}r, and Ross Girshick.
\newblock Masked autoencoders are scalable vision learners.
\newblock In \emph{Proceedings of the IEEE/CVF conference on computer vision and pattern recognition}, pages 16000--16009, 2022.

\bibitem[Kollias(2022)]{kollias2022abaw}
Dimitrios Kollias.
\newblock Abaw: Valence-arousal estimation, expression recognition, action unit detection \& multi-task learning challenges.
\newblock In \emph{Proceedings of the IEEE/CVF Conference on Computer Vision and Pattern Recognition}, pages 2328--2336, 2022.

\bibitem[Kollias(2023{\natexlab{a}})]{kollias2023abaw}
Dimitrios Kollias.
\newblock Abaw: learning from synthetic data \& multi-task learning challenges.
\newblock In \emph{European Conference on Computer Vision}, pages 157--172. Springer, 2023{\natexlab{a}}.

\bibitem[Kollias(2023{\natexlab{b}})]{kollias2023multi}
Dimitrios Kollias.
\newblock Multi-label compound expression recognition: C-expr database \& network.
\newblock In \emph{Proceedings of the IEEE/CVF Conference on Computer Vision and Pattern Recognition}, pages 5589--5598, 2023{\natexlab{b}}.

\bibitem[Kollias and Zafeiriou(2018)]{kollias2018aff}
Dimitrios Kollias and Stefanos Zafeiriou.
\newblock Aff-wild2: Extending the aff-wild database for affect recognition.
\newblock \emph{arXiv preprint arXiv:1811.07770}, 2018.

\bibitem[Kollias and Zafeiriou(2019)]{kollias2019expression}
Dimitrios Kollias and Stefanos Zafeiriou.
\newblock Expression, affect, action unit recognition: Aff-wild2, multi-task learning and arcface.
\newblock \emph{arXiv preprint arXiv:1910.04855}, 2019.

\bibitem[Kollias and Zafeiriou(2021{\natexlab{a}})]{kollias2021affect}
Dimitrios Kollias and Stefanos Zafeiriou.
\newblock Affect analysis in-the-wild: Valence-arousal, expressions, action units and a unified framework.
\newblock \emph{arXiv preprint arXiv:2103.15792}, 2021{\natexlab{a}}.

\bibitem[Kollias and Zafeiriou(2021{\natexlab{b}})]{kollias2021analysing}
Dimitrios Kollias and Stefanos Zafeiriou.
\newblock Analysing affective behavior in the second abaw2 competition.
\newblock In \emph{Proceedings of the IEEE/CVF International Conference on Computer Vision}, pages 3652--3660, 2021{\natexlab{b}}.

\bibitem[Kollias et~al.()Kollias, Schulc, Hajiyev, and Zafeiriou]{kollias2020analysing}
D Kollias, A Schulc, E Hajiyev, and S Zafeiriou.
\newblock Analysing affective behavior in the first abaw 2020 competition.
\newblock In \emph{2020 15th IEEE International Conference on Automatic Face and Gesture Recognition (FG 2020)(FG)}, pages 794--800.

\bibitem[Kollias et~al.(2019{\natexlab{a}})Kollias, Sharmanska, and Zafeiriou]{kollias2019face}
Dimitrios Kollias, Viktoriia Sharmanska, and Stefanos Zafeiriou.
\newblock Face behavior a la carte: Expressions, affect and action units in a single network.
\newblock \emph{arXiv preprint arXiv:1910.11111}, 2019{\natexlab{a}}.

\bibitem[Kollias et~al.(2019{\natexlab{b}})Kollias, Tzirakis, Nicolaou, Papaioannou, Zhao, Schuller, Kotsia, and Zafeiriou]{kollias2019deep}
Dimitrios Kollias, Panagiotis Tzirakis, Mihalis~A Nicolaou, Athanasios Papaioannou, Guoying Zhao, Bj{\"o}rn Schuller, Irene Kotsia, and Stefanos Zafeiriou.
\newblock Deep affect prediction in-the-wild: Aff-wild database and challenge, deep architectures, and beyond.
\newblock \emph{International Journal of Computer Vision}, pages 1--23, 2019{\natexlab{b}}.

\bibitem[Kollias et~al.(2021)Kollias, Sharmanska, and Zafeiriou]{kollias2021distribution}
Dimitrios Kollias, Viktoriia Sharmanska, and Stefanos Zafeiriou.
\newblock Distribution matching for heterogeneous multi-task learning: a large-scale face study.
\newblock \emph{arXiv preprint arXiv:2105.03790}, 2021.

\bibitem[Kollias et~al.(2023)Kollias, Tzirakis, Baird, Cowen, and Zafeiriou]{kollias2023abaw2}
Dimitrios Kollias, Panagiotis Tzirakis, Alice Baird, Alan Cowen, and Stefanos Zafeiriou.
\newblock Abaw: Valence-arousal estimation, expression recognition, action unit detection \& emotional reaction intensity estimation challenges.
\newblock In \emph{Proceedings of the IEEE/CVF Conference on Computer Vision and Pattern Recognition}, pages 5888--5897, 2023.

\bibitem[Kollias et~al.(2024{\natexlab{a}})Kollias, Sharmanska, and Zafeiriou]{kollias2024distribution}
Dimitrios Kollias, Viktoriia Sharmanska, and Stefanos Zafeiriou.
\newblock Distribution matching for multi-task learning of classification tasks: a large-scale study on faces \& beyond.
\newblock In \emph{Proceedings of the AAAI Conference on Artificial Intelligence}, pages 2813--2821, 2024{\natexlab{a}}.

\bibitem[Kollias et~al.(2024{\natexlab{b}})Kollias, Tzirakis, Cowen, Zafeiriou, Kotsia, Baird, Gagne, Shao, and Hu]{abaw6}
Dimitrios Kollias, Panagiotis Tzirakis, Alan Cowen, Stefanos Zafeiriou, Irene Kotsia, Alice Baird, Chris Gagne, Chunchang Shao, and Guanyu Hu.
\newblock The 6th affective behavior analysis in-the-wild (abaw) competition.
\newblock In \emph{Proceedings of the IEEE/CVF Conference on Computer Vision and Pattern Recognition}, pages 4587--4598, 2024{\natexlab{b}}.

\bibitem[Kollias et~al.(2024{\natexlab{c}})Kollias, Zafeiriou, Kotsia, Dhall, Ghosh, Shao, and Hu]{abaw7}
Dimitrios Kollias, Stefanos Zafeiriou, Irene Kotsia, Abhinav Dhall, Shreya Ghosh, Chunchang Shao, and Guanyu Hu.
\newblock 7th abaw competition: Multi-task learning and compound expression recognition.
\newblock \emph{arXiv preprint arXiv:2407.03835}, 2024{\natexlab{c}}.

\bibitem[Kollias et~al.(2025)Kollias, Tzirakis, Cowen, Zafeiriou, Kotsia, Granger, Pedersoli, Bacon, Baird, Gagne, et~al.]{baseline2025}
Dimitrios Kollias, Panagiotis Tzirakis, AS Cowen, S Zafeiriou, I Kotsia, Eric Granger, Marco Pedersoli, SL Bacon, Alice Baird, C Gagne, et~al.
\newblock Advancements in affective and behavior analysis: The 8th abaw workshop and competition.
\newblock 2025.

\bibitem[Lee et~al.(2023)Lee, Lee, and Yoo]{lee2023latent}
Isack Lee, Eungi Lee, and Seok~Bong Yoo.
\newblock Latent-ofer: Detect, mask, and reconstruct with latent vectors for occluded facial expression recognition.
\newblock In \emph{Proceedings of the IEEE/CVF International Conference on Computer Vision}, pages 1536--1546, 2023.

\bibitem[Lee et~al.(2025)Lee, Choi, Kim, Kim, and Nam]{NLA}
JunGyu Lee, Yeji Choi, Haksub Kim, Ig-Jae Kim, and Gi~Pyo Nam.
\newblock Navigating label ambiguity for facial expression recognition in the wild.
\newblock \emph{arXiv preprint arXiv:2502.09993}, 2025.

\bibitem[Li et~al.(2017)Li, Deng, and Du]{rafdb}
Shan Li, Weihong Deng, and JunPing Du.
\newblock Reliable crowdsourcing and deep locality-preserving learning for expression recognition in the wild.
\newblock In \emph{Proceedings of the IEEE conference on computer vision and pattern recognition}, pages 2852--2861, 2017.

\bibitem[Lin et~al.(2024)Lin, Papabathini, Wang, and Hu]{M2-Lab-Purdue2024}
Li Lin, Sarah Papabathini, Xin Wang, and Shu Hu.
\newblock Robust light-weight facial affective behavior recognition with clip.
\newblock In \emph{2024 IEEE 7th International Conference on Multimedia Information Processing and Retrieval (MIPR)}, pages 607--611. IEEE, 2024.

\bibitem[Liu et~al.(2020)Liu, Ouyang, Xu, Zhou, He, and Wen]{SAANet}
Daizong Liu, Xi Ouyang, Shuangjie Xu, Pan Zhou, Kun He, and Shiping Wen.
\newblock Saanet: Siamese action-units attention network for improving dynamic facial expression recognition.
\newblock \emph{Neurocomputing}, 413:\penalty0 145--157, 2020.

\bibitem[Liu et~al.(2021)Liu, Lin, Cao, Hu, Wei, Zhang, Lin, and Guo]{liu2021swin}
Ze Liu, Yutong Lin, Yue Cao, Han Hu, Yixuan Wei, Zheng Zhang, Stephen Lin, and Baining Guo.
\newblock Swin transformer: Hierarchical vision transformer using shifted windows.
\newblock In \emph{Proceedings of the IEEE/CVF international conference on computer vision}, pages 10012--10022, 2021.

\bibitem[Loshchilov and Hutter(2017)]{loshchilov2017decoupled}
Ilya Loshchilov and Frank Hutter.
\newblock Decoupled weight decay regularization.
\newblock \emph{arXiv preprint arXiv:1711.05101}, 2017.

\bibitem[Lucey et~al.(2010)Lucey, Cohn, Kanade, Saragih, Ambadar, and Matthews]{ckplus}
Patrick Lucey, Jeffrey~F Cohn, Takeo Kanade, Jason Saragih, Zara Ambadar, and Iain Matthews.
\newblock The extended cohn-kanade dataset (ck+): A complete dataset for action unit and emotion-specified expression.
\newblock In \emph{2010 ieee computer society conference on computer vision and pattern recognition-workshops}, pages 94--101. IEEE, 2010.

\bibitem[Lyons et~al.(1998)Lyons, Akamatsu, Kamachi, and Gyoba]{lyons1998coding_JAFFE}
Michael Lyons, Shigeru Akamatsu, Miyuki Kamachi, and Jiro Gyoba.
\newblock Coding facial expressions with gabor wavelets.
\newblock In \emph{Proceedings Third IEEE international conference on automatic face and gesture recognition}, pages 200--205. IEEE, 1998.

\bibitem[Mollahosseini et~al.(2017)Mollahosseini, Hasani, and Mahoor]{affectnet}
Ali Mollahosseini, Behzad Hasani, and Mohammad~H Mahoor.
\newblock Affectnet: A database for facial expression, valence, and arousal computing in the wild.
\newblock \emph{IEEE Transactions on Affective Computing}, 10\penalty0 (1):\penalty0 18--31, 2017.

\bibitem[Pantic et~al.(2005)Pantic, Valstar, Rademaker, and Maat]{pantic2005web_mimi}
Maja Pantic, Michel Valstar, Ron Rademaker, and Ludo Maat.
\newblock Web-based database for facial expression analysis.
\newblock In \emph{2005 IEEE international conference on multimedia and Expo}, pages 5--pp. IEEE, 2005.

\bibitem[Savchenko(2024)]{hsemotion2024}
Andrey~V Savchenko.
\newblock Hsemotion team at the 6th abaw competition: Facial expressions, valence-arousal and emotion intensity prediction.
\newblock \emph{arXiv preprint arXiv:2403.11590}, 2024.

\bibitem[She et~al.(2021)She, Hu, Shi, Wang, Shen, and Mei]{DMUE}
Jiahui She, Yibo Hu, Hailin Shi, Jun Wang, Qiu Shen, and Tao Mei.
\newblock Dive into ambiguity: Latent distribution mining and pairwise uncertainty estimation for facial expression recognition.
\newblock In \emph{Proceedings of the IEEE/CVF conference on computer vision and pattern recognition}, pages 6248--6257, 2021.

\bibitem[Tong et~al.(2022)Tong, Song, Wang, and Wang]{tong2022videomae}
Zhan Tong, Yibing Song, Jue Wang, and Limin Wang.
\newblock Videomae: Masked autoencoders are data-efficient learners for self-supervised video pre-training.
\newblock \emph{Advances in neural information processing systems}, 35:\penalty0 10078--10093, 2022.

\bibitem[Wang et~al.(2020{\natexlab{a}})Wang, Peng, Yang, Lu, and Qiao]{SCN}
Kai Wang, Xiaojiang Peng, Jianfei Yang, Shijian Lu, and Yu Qiao.
\newblock Suppressing uncertainties for large-scale facial expression recognition.
\newblock In \emph{Proceedings of the IEEE/CVF conference on computer vision and pattern recognition}, pages 6897--6906, 2020{\natexlab{a}}.

\bibitem[Wang et~al.(2020{\natexlab{b}})Wang, Peng, Yang, Meng, and Qiao]{wang2020region}
Kai Wang, Xiaojiang Peng, Jianfei Yang, Debin Meng, and Yu Qiao.
\newblock Region attention networks for pose and occlusion robust facial expression recognition.
\newblock \emph{IEEE Transactions on Image Processing}, 29:\penalty0 4057--4069, 2020{\natexlab{b}}.

\bibitem[Wu and Cui(2023)]{LANet}
Zhiyu Wu and Jinshi Cui.
\newblock La-net: Landmark-aware learning for reliable facial expression recognition under label noise.
\newblock In \emph{Proceedings of the IEEE/CVF International Conference on Computer Vision}, pages 20698--20707, 2023.

\bibitem[Yu et~al.(2024)Yu, Wei, Cai, Zhao, Zhang, Wang, Xie, Zhu, Zhu, Liu, et~al.]{USTC-IAT-United2024}
Jun Yu, Zhihong Wei, Zhongpeng Cai, Gongpeng Zhao, Zerui Zhang, Yongqi Wang, Guochen Xie, Jichao Zhu, Wangyuan Zhu, Qingsong Liu, et~al.
\newblock Exploring facial expression recognition through semi-supervised pre-training and temporal modeling.
\newblock In \emph{Proceedings of the IEEE/CVF Conference on Computer Vision and Pattern Recognition}, pages 4880--4887, 2024.

\bibitem[Zafeiriou et~al.(2017)Zafeiriou, Kollias, Nicolaou, Papaioannou, Zhao, and Kotsia]{zafeiriou2017aff}
Stefanos Zafeiriou, Dimitrios Kollias, Mihalis~A Nicolaou, Athanasios Papaioannou, Guoying Zhao, and Irene Kotsia.
\newblock Aff-wild: Valence and arousal ‘in-the-wild’challenge.
\newblock In \emph{Computer Vision and Pattern Recognition Workshops (CVPRW), 2017 IEEE Conference on}, pages 1980--1987. IEEE, 2017.

\bibitem[Zeng et~al.(2022)Zeng, Lin, Yan, Liu, Wang, and Tang]{face2exp}
Dan Zeng, Zhiyuan Lin, Xiao Yan, Yuting Liu, Fei Wang, and Bo Tang.
\newblock Face2exp: Combating data biases for facial expression recognition.
\newblock In \emph{Proceedings of the IEEE/CVF Conference on Computer Vision and Pattern Recognition}, pages 20291--20300, 2022.

\bibitem[Zeng et~al.(2018)Zeng, Shan, and Chen]{IPA2LT}
Jiabei Zeng, Shiguang Shan, and Xilin Chen.
\newblock Facial expression recognition with inconsistently annotated datasets.
\newblock In \emph{Proceedings of the European conference on computer vision (ECCV)}, pages 222--237, 2018.

\bibitem[Zhang et~al.(2024{\natexlab{a}})Zhang, Qiu, Liu, Li, Du, Guo, and Yu]{netease2024}
Wei Zhang, Feng Qiu, Chen Liu, Lincheng Li, Heming Du, Tiancheng Guo, and Xin Yu.
\newblock Affective behaviour analysis via integrating multi-modal knowledge.
\newblock \emph{arXiv preprint arXiv:2403.10825}, 2024{\natexlab{a}}.

\bibitem[Zhang et~al.(2021)Zhang, Wang, and Deng]{RUL}
Yuhang Zhang, Chengrui Wang, and Weihong Deng.
\newblock Relative uncertainty learning for facial expression recognition.
\newblock \emph{Advances in Neural Information Processing Systems}, 34:\penalty0 17616--17627, 2021.

\bibitem[Zhang et~al.(2022{\natexlab{a}})Zhang, Wang, Ling, and Deng]{EAC}
Yuhang Zhang, Chengrui Wang, Xu Ling, and Weihong Deng.
\newblock Learn from all: Erasing attention consistency for noisy label facial expression recognition.
\newblock In \emph{European Conference on Computer Vision}, pages 418--434. Springer, 2022{\natexlab{a}}.

\bibitem[Zhang et~al.(2024{\natexlab{b}})Zhang, Li, Liu, Deng, et~al.]{MEK}
Yuhang Zhang, Yaqi Li, Xuannan Liu, Weihong Deng, et~al.
\newblock Leave no stone unturned: Mine extra knowledge for imbalanced facial expression recognition.
\newblock \emph{Advances in Neural Information Processing Systems}, 36, 2024{\natexlab{b}}.

\bibitem[Zhang et~al.(2022{\natexlab{b}})Zhang, Sun, Li, and Wang]{zhang2022man}
Ziyang Zhang, Xiao Sun, Jia Li, and Meng Wang.
\newblock Man: Mining ambiguity and noise for facial expression recognition in the wild.
\newblock \emph{Pattern Recognition Letters}, 164:\penalty0 23--29, 2022{\natexlab{b}}.

\bibitem[Zhou et~al.(2016)Zhou, Khosla, Lapedriza, Oliva, and Torralba]{zhou2016learning_CAM}
Bolei Zhou, Aditya Khosla, Agata Lapedriza, Aude Oliva, and Antonio Torralba.
\newblock Learning deep features for discriminative localization.
\newblock In \emph{Proceedings of the IEEE conference on computer vision and pattern recognition}, pages 2921--2929, 2016.

\bibitem[Zhou et~al.(2024)Zhou, Lu, Ling, Wang, and Liu]{CtyunAI2024}
Weiwei Zhou, Jiada Lu, Chenkun Ling, Weifeng Wang, and Shaowei Liu.
\newblock Boosting continuous emotion recognition with self-pretraining using masked autoencoders, temporal convolutional networks, and transformers.
\newblock \emph{arXiv preprint arXiv:2403.11440}, 2024.

\end{thebibliography}
